\documentclass{article}
\usepackage{spconf,amsmath,graphicx}
\usepackage{multirow, amssymb}
\usepackage{array}
\usepackage{caption, subcaption, tabu, epsfig}
\usepackage{booktabs}

\makeatletter
\newcommand{\thickhline}{%
    \noalign {\ifnum 0=`}\fi \hrule height 1pt
    \futurelet \reserved@a \@xhline
}
\newcommand\blfootnote[1]{%
  \begingroup
  \renewcommand\thefootnote{}\footnote{#1}%
  \addtocounter{footnote}{-1}%
  \endgroup
}

\title{QUERY-BY-EXAMPLE ON-DEVICE KEYWORD SPOTTING}
\name{Byeonggeun Kim, Mingu Lee, Jinkyu Lee, Yeonseok Kim, and Kyuwoong Hwang\thanks{Qualcomm AI Research is an initiative of Qualcomm Technologies, Inc}}
\address{Qualcomm AI Research\\
Hakdong-ro, Gangnam-gu, Seoul, Republic of Korea}
\begin{document}
%
\maketitle
\begin{abstract}
A keyword spotting (KWS) system determines the existence of, usually predefined, keyword in a continuous speech stream. This paper presents a query-by-example on-device KWS system which is user-specific. The proposed system consists of two main steps: query enrollment and testing. In query enrollment step, phonetic posteriors are output by a small-footprint automatic speech recognition model based on connectionist temporal classification. Using the phonetic-level posteriorgram, hypothesis graph of finite-state transducer (FST) is built, thus can enroll any keywords thus avoiding an out-of-vocabulary problem. In testing, a log-likelihood is scored for input audio using the FST. We propose a threshold prediction method while using the user-specific keyword hypothesis only. The system generates query-specific negatives by rearranging each query utterance in waveform. The threshold is decided based on the enrollment queries and generated negatives. We tested two keywords in English, and the proposed work shows promising performance while preserving simplicity.
\end{abstract}
\section{Introduction}
\label{sec:intro}

Keyword spotting (KWS) has widely been used in personal devices like mobile phones and home appliances for detecting keywords which are usually compounded of one or two words. The goal is to detect the keywords from real-time audio stream. For practical use, it is required to achieve low false rejection rate (FRR) while keeping low false alarms (FAs) per hour.

Many previous works consider predefined keywords to reach promising performance. Keywords such as ``Alexa", ``Okay/Hey Google", ``Hey Siri" and ``Xiaovi Xiaovi" are the examples. They collect numerous variations of a specific keyword utterance and train neural networks (NNs) which have been promising method in the field.  \cite{Guo2018, Chen2018} have acoustic encoder and sequence matching decoder as separate modules. The NN-based acoustic models (AMs) predict senone-level posteriors. Sequence matching, traditionally modeled by hidden Markov models (HMMs), interprets the AM outputs into keyword and background parts. Meanwhile, \cite{Shan2018, wangadversarial, coucke2018, Alvarez} have end-to-end NN architectures to directly determine the presence of keywords. They use recurrent neural networks (RNN) with attention layers \cite{Shan2018, wangadversarial}, dilated convolution network \cite{coucke2018}, or filters based on singular value decomposition \cite{Alvarez}.

On the other hand, there have been query-by-example approaches which detect query keywords of any kinds. Early approaches use automatic speech recognition (ASR) phonetic posterior as a posteriorgram and exploit dynamic time warping (DTW) to compare keyword samples and test utterances \cite{hazen2009query, zhang2009unsupervised, anguera2013memory}. \cite{zhuang2016unrestricted, lugosch2018donut} also used posteriorgram while using connectionist temporal classification (CTC) ASR. \cite{zhuang2016unrestricted} used an edit distance metric, and \cite{lugosch2018donut} directly used posteriors of ASR. Furthermore, \cite{Chen2015} computes a simple similarity scores of LSTM output vectors between enrollment and test utterance. Recently, end-to-end NN based query-by-example systems are suggested \cite{he2017streaming, audhkhasi2017end}. \cite{he2017streaming} uses a recurrent neural network transducer (RNN-T) model biased with attention over keyword. \cite{audhkhasi2017end} suggests to use text query instead of audio.

Meanwhile, there have been other groups who explored keyword spotting problem. \cite{Myer2018, Tang2018, Pandey2018, fernandez2007application} solve multiple keyword detection. \cite{Menon2018, Chen2018-2} focus on KWS tasks with small dataset. \cite{Menon2018} use DTW to augment the data, and \cite{Chen2018-2} suggests a few-shot meta-learning approach. 

In this paper, we propose a simple yet powerful query-by-example on-device KWS approach using user-specific queries. Our system provides user-specific model by utilizing a few keyword utterances spoken by a single user. The system uses posteriorgram based graph matching algorithm using a small-footprint ASR. An ASR based CTC \cite{gravesCTC} outputs phonetic posteriors, and we build a hypothesis graph of finite-state transducer (FST). The posteriorgram consists of phonetic output which frees the model from out-of-vocabulary problem. On testing, the system determines whether an input audio contains the keyword or not through a log-likelihood score according to the graph which includes constraints of phonetic hypothesis. Despite of the score normalization, score-based query-by-example on-device KWS systems usually suffer from threshold decision, because there are not enough negative examples in on-device system. We predict user-specific threshold by keyword hypothesis graphs. We generate query-specific negatives by rearranging positives in waveform. Then we predicts a threshold by using positives and generated negatives. While keeping this simplicity, our approach shows comparable performances with recent KWS systems.

The rest of the paper is organized as follows. In Section 2, the KWS system is described including the acoustic model, the FST in the decoder, and the threshold prediction method. The performance evaluation results are discussed in Section 3 followed by the conclusion in Section 4.

\section{Query-by-example KWS system}
\label{sec:format}

Our system consists of three parts, acoustic model, decoder, and threshold prediction part. In subsections, we denote acoustic model input features as $X = x_1, x_2, \cdots, x_T$ where $x_t \in \mathbb{R}^M$ and $t$ is a time frame index. Corresponding labels are $Y=y_{1}, y_{2}, \cdots, y_{K}$ and usually $K < T$.

\subsection{Acoustic model}

We exploit a CTC acoustic model \cite{gravesCTC}. We denote activation of ASR as $O=o_1, o_2, \cdots, o_T$ where $o_t\in \mathbb{R}^N$ and let $o^n_t$ as activation of unit n at time t. Thus $o^n_t$ is a probability of observing n at time t. CTC uses a extra blank output $\phi$. We denote $L'=L\cup \{\phi, space\}$ where $L$ is the set of 39 context-independent phonemes. The space output implies a short pause between words. We let $L'(T)$ as sequence set of length T, where their elements are in $L'$. Then, conditional probability of path $P$ given $X$ is $p(P|X) = \prod_{t=1}^{T}p(o^{P_t}_t)$ where $\forall P \in L'(T)$.

\cite{gravesCTC} suggests a many-to-one mapping $B$ which maps activation $O$ to label sequence $Y$. The mapping collapses repeats and removes blank output $\phi$, e.g. $B(x\phi yy \phi z)=B(x\phi\phi yz \phi)=xyz$. The conditional probability $P(Y|X)$ is marginalizing of possible paths for Y and is defined as,

\begin{equation}\label{eq:CTC}
    p(Y|X) = \sum_{P\in B^{-1}(Y)}{p(P|X)}.
\end{equation}

\subsection{Keyword spotting decoder}

The keyword spotting decoder operates in two phases: an enrollment step and testing. In the enrollment step, using AM output of the query utterance, the model finds the hypothesis and build FSTs for the path. While testing, the model calculates the score and determines whether the input utterance contains the keyword using the hypothesis.

\subsubsection{Query enrollment}

In the enrollment step, the system uses a few clean utterances of a keyword spoken by a single user. We use simple and heuristic method, max-decoding. We follow the component of maximum-posterior at each time frame. For each time step $t$, we choose $\underset{n}{\text{argmax}}(o_{t}^n, n=1, \cdots, N)$ and get a path P. The hypothesis is defined by the mapping $B$, as $B(P)$.

A keyword `Hey Snapdragon' gives a hypothesis like `HH.EY. .S.N.AE.P.T. .A.AE.G.AH.N'. With the hypothesis as a sequential phonetic constraint, we generate left-to-right FST systems.

\begin{figure}[t]
  \centering
  \includegraphics[width=\linewidth]{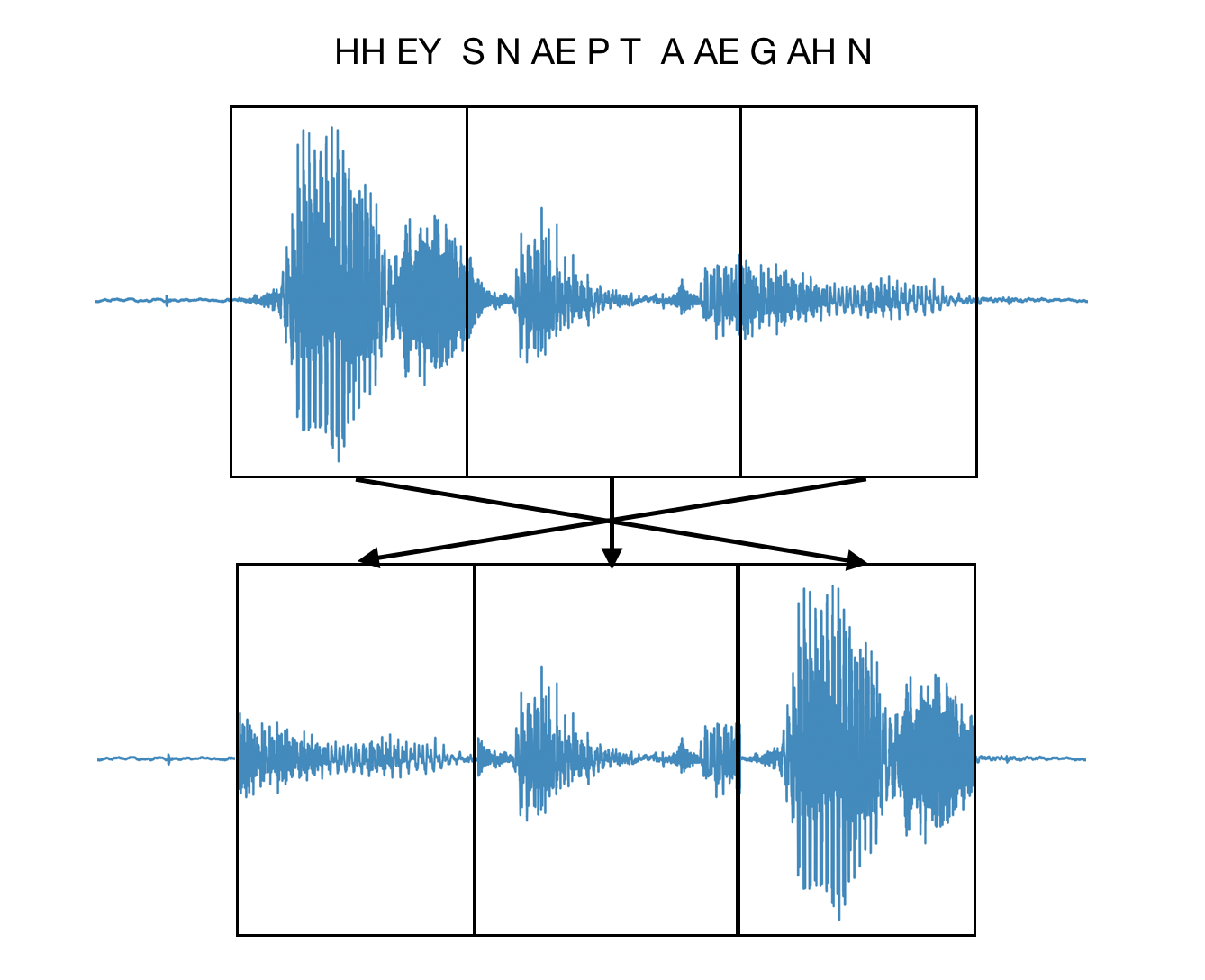}
  \vspace*{-20pt}
  \caption{Example of a generated negative from a query utterance, `Hey Snapdragon'. The query utterance is divided into three in waveform and shuffled.}
  \vspace*{-15pt}
  \label{fig:neg_gen}
\end{figure}

\subsubsection{Keyword spotting}

In testing, the system calculates a score of a test utterance for hypothesis FSTs. Assume that the FST has $L$ distinct possible states $S=[s^{(i)}],\:i=1,2,\cdots, L$ where $s^{(\phi)}$ denotes the blank state. The FST is left-to-right, therefore, has an ordered label Hypothesis $Y' = y'_1, y'_2, \cdots, y'_K$ where $y'_k\in S, \forall k$. Given the hypothesis, the score is log likelihood of a test input, $X' = x'_1, x'_2, \cdots, x'_T$.

At time step $t$, the activation of AM is $o_t$ and we denote the corresponding FST state as $q_t\in S$. The transition probability $a_{ij}$ is $p(q_t=s^{(j)}|q_{t-1}=s^{(i)})$. The hypothesis limits the transition probability as Eq.(\ref{eq:transprob}), where $q_{t-1} =y'_{l-1}$. If $q_t=s_\phi$, then $q_t=q_{t-1}$, i.e. remaining in the previous state.  Hypothesis $Y'$ is usually shorter than $X'$ because we use the mapping $B$ to get $Y'$. Therefore it is more likely to remain at a current state than moving to the next. We naively choose the transition probabilities to reflect the scenario.

\begin{equation}\label{eq:transprob}
    a_{ij}(t) = 
    \begin{cases}
    1/3, & \text{if } q_t \in \{y'_l,y'_{l-1},s^{(\phi)}\} \\
    0, & \text{otherwise.} \\
    \end{cases}
\end{equation}

\blfootnote{Snapdragon is a registered trademark of Qualcomm Incorporated.}

A log likelihood is,

\begin{multline}\label{eq:score}
    \log p(X'|Y') = \log\{\sum_{q}p(q|Y')p(X'|Y', q)\}\\
    \approx \underset{q, t_0} {\max}[\log\{\pi\prod_{t=t_0+1}^{T}{a_{q_{t-1}q_{t}}}\prod_{t=t_0}^{T}{\frac{p(q_t|x'_t)p(x'_t)}{p(q_t)}}\}] \\
    \propto \underset{q, t_0} {\max}[\log\{\pi\prod_{t=t_0+1}^{T}{a_{q_{t-1}q_{t}}}\prod_{t=t_0}^{T}{p(q_t|x'_t)}\}],
\end{multline}

where $\pi$ denotes the initial state probability, and $\pi = p(q_1=y'_1)=1$ for a given path. The $p(q|Y')$ is product of transition probabilities, and the likelihood, $p(X'|Y', q)$ is proportional to the posteriors of the AM. Here $p(x'_t)$ and the state prior $p(q_t)$ are assumed to be uniform.

We normalize the score by dividing Eq.(\ref{eq:score}) by the number of non-blank states, $|\{q_t|t=1,\cdots,T, q_t\neq s^{(\phi)}\}|$. We find $q$ and $t_0$ which maximize Eq.(\ref{eq:score}) by beam searching. During the search, we consider each time step $t$ as a initial time $t_0$. By doing this, the system can spot the keyword in a long audio stream.

\begin{figure}[t]
  \centering
  \includegraphics[width=\linewidth]{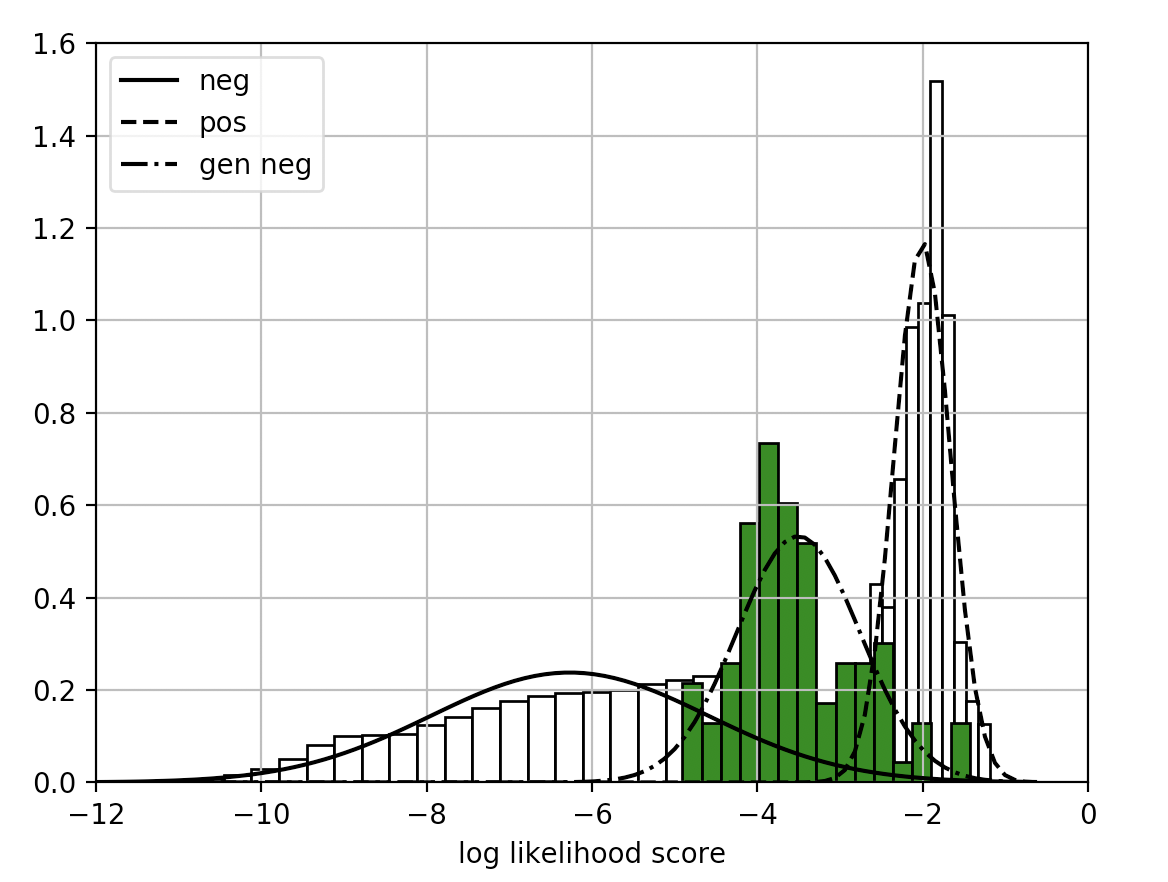}
  \vspace*{-20pt}
  \caption{A histogram of query, negative and generated negative log likelihood scores for hypothesis FSTs of a single speaker. Colored histogram shows generated negatives.}
  \vspace*{-20pt}
  \label{fig:pdf}
\end{figure}

\subsection{On-device threshold prediction}

In this section, query set is $Q = \{X'_1, X'_2, \cdots, X'_A\}$, and corresponding hypothesis set is $H = \{Y'_1, Y'_2, \cdots, Y'_A\}$. $F_Y(X)$ is a mapping from a test utterance, X, to log likelihood score for a hypothesis $Y$. We denote negative utterances as $Z_1, Z_2, \cdots, Z_B$. The hypothesis computes positive scores from each other's query. A threshold $\delta$ is defined as,

\begin{equation}
\begin{split}
\label{eq:tr1}
    \delta_{(Q, H)} = {\tau \over A(A-1)} \sum_{(a, a')} {F_{Y'_a}(X'_{a'})|_{a\neq a'}} \\
    + {(1-\tau)\over A\cdot B}\sum_{(a,b)}{F_{Y'_a}(Z_b)}
\end{split}
\end{equation}

where $\tau$ is a hyperparameter in $[0, 1]$, $a, a' \in [A]$ and $b \in [B]$. Eq.(\ref{eq:tr1}) means the threshold as a score between mean of positive scores and that of negative scores.

We generate query-specific negatives from queries. Figure~\ref{fig:neg_gen} shows an example of a keyword, 'Hey Snapdragon'. Each positive is divided to sub-parts and shuffled in waveform. We overlap 16 samples of each part boundary and apply them one-sided triangular windows to guarantee smooth waveform transition and to prevent undesirable discontinuities, i.e. impulsive noises. Figure~\ref{fig:pdf} plots an example of histograms of queries, negatives, and generated negatives of hypothesis FSTs from a single speaker. A probability distribution is drawn in histogram while assuming Gaussian distribution for better visualization. We used the generated negatives as $\{Z_b\}$.

\begin{figure}[t]
  \centering
  \includegraphics[width=\linewidth]{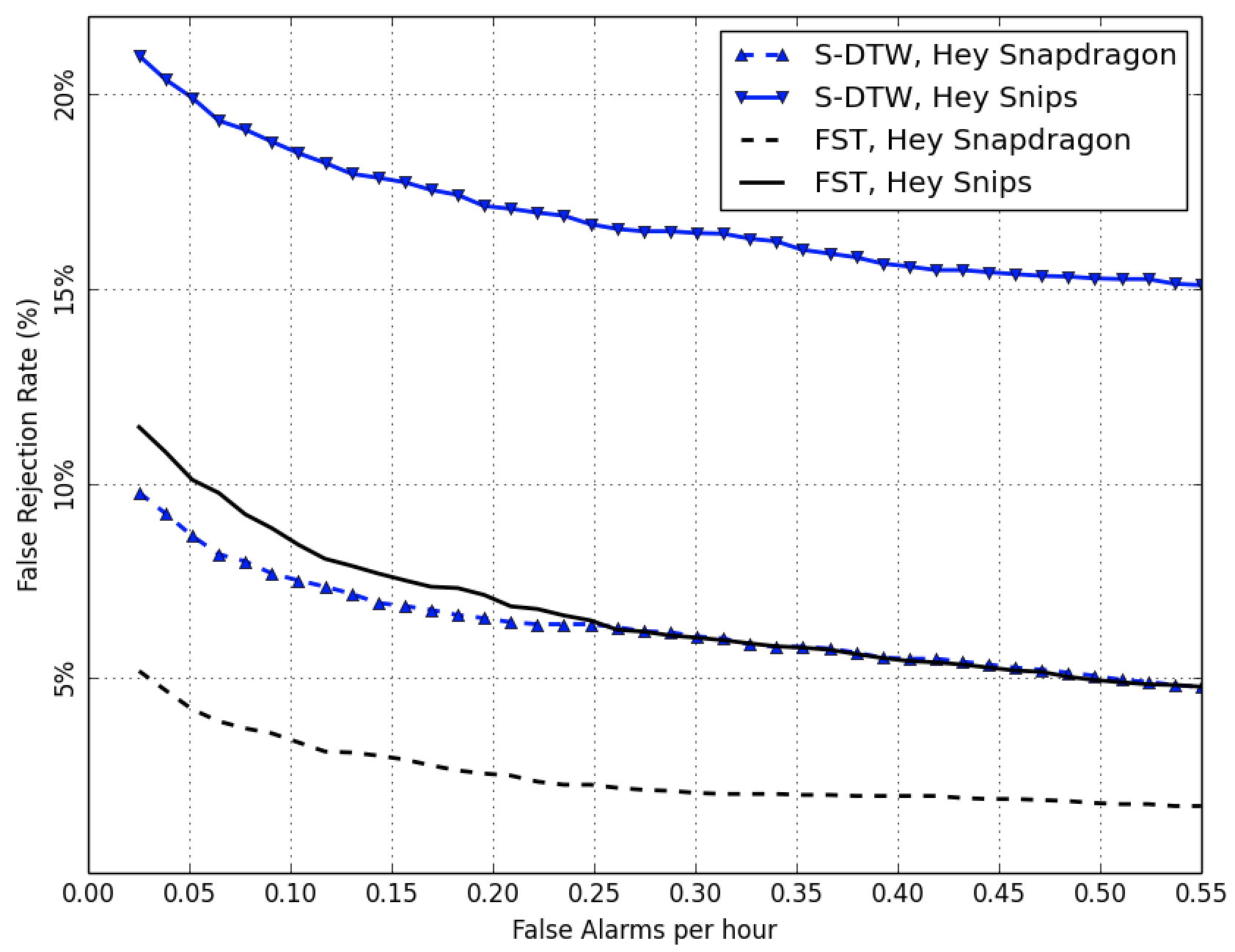}
  \vspace*{-10pt}
  \caption{Comparison of baseline, the S-DTW with the FST constrained by phonetic hypothesis.}
  \vspace*{-15pt}
  \label{fig:DETcurve}
\end{figure}

\section{Experiments}
\label{sec:pagestyle}

\subsection{Experimental setup}

\subsubsection{Query and testing data}

Many previous works experiment with their own data which are not accessible. In some literature, only relative performances are reported, thus the results are hard to compare with each other and are not reproducible. To be free from this issue, we use public and well-known data.

\footnotetext[1]{https://developer.qualcomm.com/project/keyword-speech-dataset}

\begin{table*}[]
\caption{FRR (\%) at 0.05 FAs per hour for clean and SNR levels \{10 dB, 6 dB, 0 dB\} of positives.}
\vspace*{-5pt}
\label{comparison to baseline}
\centering
\begin{tabular}{cl|llll|l}
\thickhline
Method                 & \multicolumn{1}{c|}{Keyword} & \multicolumn{1}{c}{clean} & \multicolumn{1}{c}{10 dB} & \multicolumn{1}{c}{6 dB} & \multicolumn{1}{c|}{0 dB} & \multicolumn{1}{c}{Avg.} \\ \hline
\multirow{2}{*}{S-DTW} & Hey Snapdragon               & 1.35                      & 3.84                      & 8.01                     & 21.6                      & 8.70                     \\
                      & Hey Snips                  & 10.5                      & 15.8                      & 20.7                     & 32.8                      & 19.9                     \\
\multirow{2}{*}{FST}   & Hey Snapdragon               & 0.53                      & 0.83                      & 3.22                     & 12.2                      & 4.19                     \\
                      & Hey Snips                    & 1.85                      & 5.36                      & 8.59                     & 24.7                      & 10.13                    \\ \thickhline
\end{tabular}
\end{table*}

\begin{table*}[]
\caption{Comparison of FRR (\%) of various KWS systems at given FAs per hour levels.}
\label{compare table}
\vspace*{-5pt}
\centering
\begin{tabular}{ll|llccc}
\thickhline
\multicolumn{1}{c}{Method} & \multicolumn{1}{c|}{Keyword} & \multicolumn{1}{c}{Params} & \multicolumn{1}{c}{SNR} & FRR @ 1 FA/ hr & FRR @ 0.5 FA/ hr & \multicolumn{1}{l}{FRR @ 0.05 FA/ hr} \\ \hline
Shan et al. {[}3{]}        & Xiao ai tong xue             & 84 k                       & \multicolumn{1}{c}{-}   & 1.02           & -                & -                                     \\
Coucke et al. {[}5{]}      & Hey snips                    & 222 k                       & 5 dB\footnotemark[2]                    & -              & 1.60             & -                                     \\
Wang et al. {[}4{]}        & Hai xiao wen                 & \multicolumn{1}{c}{-}      & \multicolumn{1}{c}{-}   & 4.17           & -                & -                                     \\
He et al. {[}13{]}             & Personal Name\footnotemark[3]               & \multicolumn{1}{c}{-}      & \multicolumn{1}{c}{-}   & -              & -                & 8.9                                   \\ \thickhline
\multirow{2}{*}{S-DTW}     & Hey Snapdragon               & \multirow{4}{*}{211 k}      & \multirow{4}{*}{6 dB}   & 3.12           & 4.46             & 8.01                                  \\ 
                          & Hey Snips                    &                            &                         & 13.30          & 15.07            & 20.69                                 \\ \cline{1-2}
\multirow{2}{*}{FST}       & Hey Snapdragon               &                            &                         & 0.62           & 1.04             & 3.22                                  \\
                          & Hey Snips                    &                            &                         & 2.79           & 3.77             & 8.58                                  \\ \thickhline
\end{tabular}
\vspace*{-10pt}
\end{table*}

We use two query keywords in English, `Hey Snapdragon' and `Hey Snips'. The audio data of `Hey Snips' is introduced at \cite{coucke2018}. We select 61 speakers who have at least 11 `Hey Snips' utterances each. We use 993 utterances from the data. `Hey Snapdragon' utterances are from a publicly available dataset\footnotemark[1]. There are 50 speakers and each of them speaks the keyword 22 or 23 times. In total, there are 1,112 `Hey Snapdragon'. At each user-specific test, 3 query utterance are randomly picked and rest are used as positive test samples. We augment the positive utterances using five types of noises, \{babble, car, music, office, typing\} at three signal-to-noise ratios (SNR) \{10 dB, 6 dB and 0 dB\}. 


We use WSJ-SI200 \cite{paul1992design} as negative samples. We sampled 24 hrs of WSJ-SI200 and segmented the whole audio stream into 2 seconds long. We augment each data with one of the five noise types, \{babble, car, music, office, typing\} and one SNR ratio among \{10 dB, 6 dB and 0 dB\}. Noise type and SNR are randomly selected.

\subsubsection{Acoustic model details}

The model is trained with Librispeech \cite{panayotov2015librispeech} data. Noises, \{babble, music, street\}, are added at uniform random SNRs in $[-3, 15]$ dB range. For more generalized model, we distorted the data by speech rate, power and reverberation. We changeed the speech rate with uniform random rates between $0.9$ and $1.2$. For reverberation, we used several measured room impulse responses in a moderately reverberant meeting rooms in an office building. From the term `power', we meant the input level augmentation for which we changed the peak amplitudes of the input waveforms to have a random value between $0$ dB and $-40$ dB in the normalized full scale.

Input features are 40-dimensional Per-channel energy normalization (PCEN) mel-filterbank energy \cite{Wang} with 30 ms window and 10 ms frame shift. The model has two convolutional layers followed by five unidirectional LSTM layers. Each covolutional layer is followed by batch normalization and activation function. Each LSTM layer has 256 LSTM cells. On top, there are a fully-connected layer and a softmax layer. Through the trade-off between ASR performance and network size, the model has 211 k number of parameters and shows 16.61 \% phoneme-error-rate (PER) and 48.04 \% word-error-rate (WER) on Librispeech test-clean dataset without prior linguistic information.

\footnotetext[2]{Coucke et al. \cite{coucke2018} augmented the positive dev and test datasets by only 5 dB, while our 6 dB is only for positive dev. Our test dataset are augmented by \{10, 6, 0\} dB.}
\footnotetext[3]{He et al. \cite{he2017streaming} used queries like 'Olivia' and 'Erica'.}

\subsection{Results}

We tested 111 user-specific KWS systems. 50 are from the query `Hey Snapdragon' and the rest are from `Hey Snips'. We used three queries from a given speaker for an enrollment. When we use one or two queries instead, the relative increase of FRR (\%) at 0.5 FA per hour are 222.05 \% or 2.47 \% respectively at 6 dB SNR. The scores from three hypothesis are averaged for each test.

\subsubsection{Baseline}

Some previous works exploit DTW to compare the query and test sample \cite{hazen2009query, zhang2009unsupervised, anguera2013memory}. We exploit DTW as our baseline while using the CTC-based AM model. We use KL-divergence as DTW distance, and allow a subsequence as an optmial path, which refers to subsequence DTW (S-DTW) \cite{muller2007dynamic}. The score is normalized by input length of DTW corresponding to the optimal path.

\subsubsection{FST constrained by phonectic hypothesis}

We build 3 hypothesis FST for each system. We tested all 111 user-specific models and average them by keywords.
Table ~\ref{comparison to baseline} compares the baseline, the S-DTW with the FST method, and we average the performances for the four SNR levels to plot a ROC curve, shown in Figure ~\ref{fig:DETcurve}. The method using FST consistently outperforms the S-DTW while using a same query, and `Hey Snapdragon' stands out than `Hey Snips'. The query word, `Hey Snips' is short and false alarms are more likely to occur. The performance is heavily influenced by the type of keyword and this result is also specified in \cite{he2017streaming}. 

\begin{figure}[t]
  \centering
  \includegraphics[width=\linewidth]{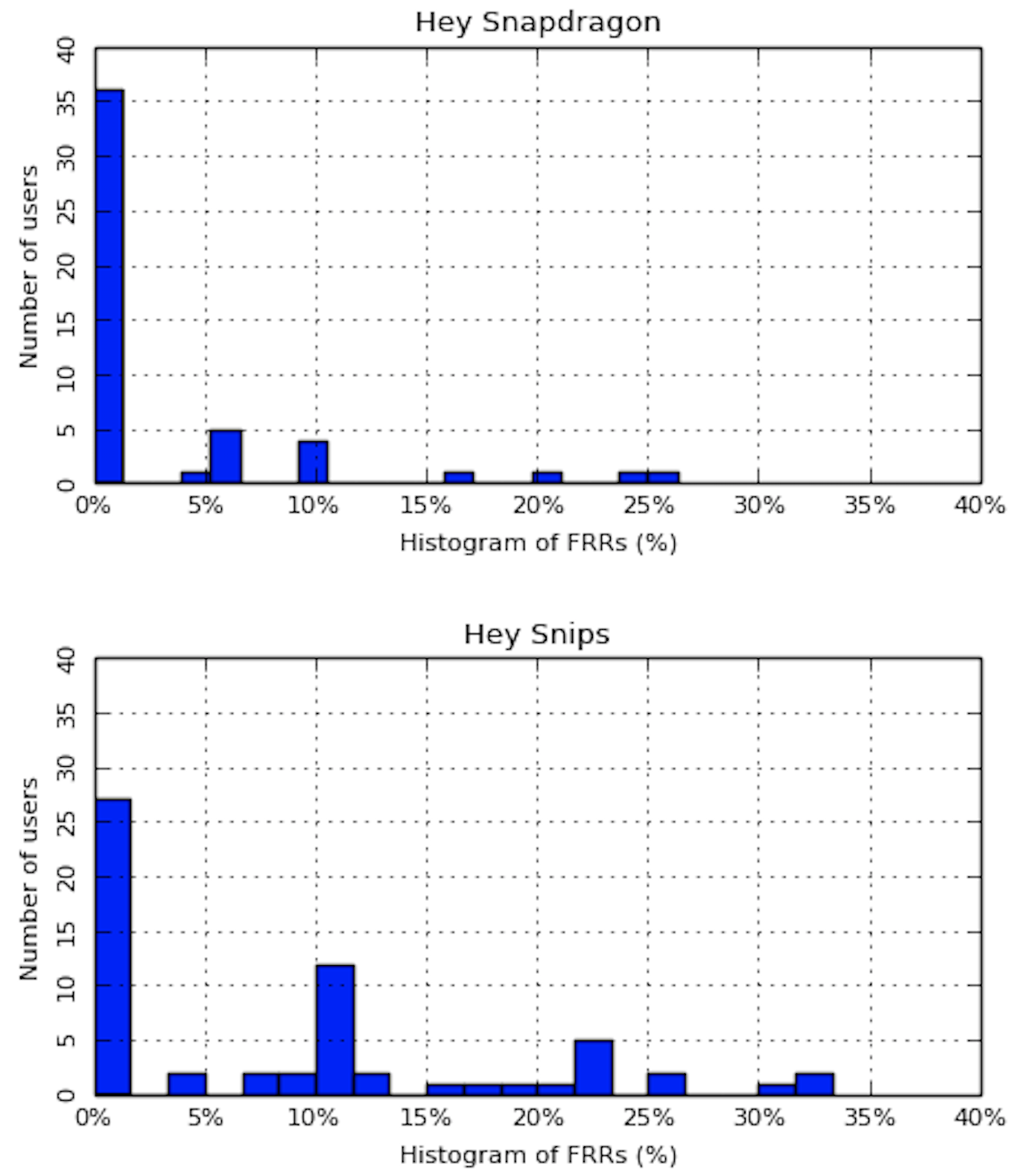}
  \vspace*{-20pt}
  \caption{Histograms of FRRs (\%) at 0.05 FA/ hr per user model.}
  \vspace*{-10pt}
  \label{fig:peruser}
\end{figure}

In Figure \ref{fig:peruser}, we plot a histogram which shows the FRR by users. Most user models show low FRR except some outliers. 

Due to the limited data access, direct result comparison with previous works became difficult. Nevertheless, we compared our results with others in Table~\ref{compare table} to show that the results are comparable to that of predefined KWS systems \cite{Shan2018, coucke2018, wangadversarial} and query-by-example system \cite{he2017streaming}. Blanks in the table implies unknown information.

\begin{figure}[t]
  \centering
  \includegraphics[width=\linewidth]{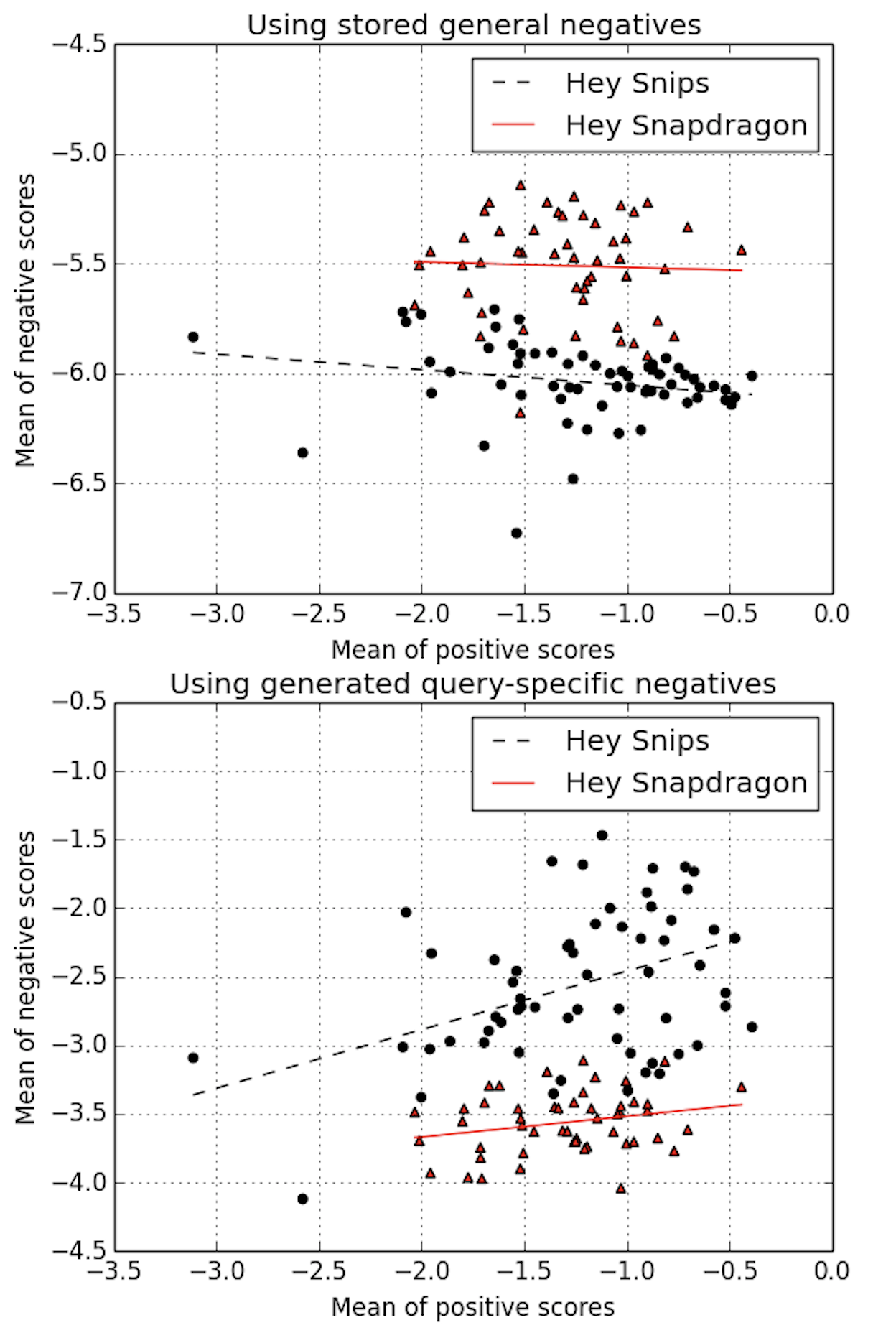}
  \vspace*{-20pt}
  \caption{Comparison of baseline with query-specific generated negatives. The graphs show relationship between the mean positives and the mean negatives and their best-fit in lines.}
  \vspace*{-10pt}
  \label{fig:correlation}
\end{figure}

\subsubsection{On-device threshold prediction}

We tested a naive threshold prediction approach as a baseline. The baseline assumes a scenario that a device stores randomly chosen 100 general negatives. 50 negatives are from clean and the rest are from augmented data mentioned in section 3.1.1. $A=3$ and $B=100$ in Eq.(\ref{eq:tr1}).

The proposed method exploits query-specific negatives. For each query, we divide the waveform into three parts with the same lengths, thus there are five ways to shuffle to make it different from the original signal. There are three queries for each enrollment and, therefore we have 15 generated negatives. Each hypothesis from a query uses other two queries as positives and their generated negatives as negatives, thus $A=3$ and $B=10$.

Figure \ref{fig:correlation} shows mean of positive and that of negative scores for 111 user-specific models. The baseline shows low and even negative correlation coefficient (R) value. R values for `Hey Snapdragon' and `Hey Snips' are -0.04 and -0.21 respectively. Meanwhile, the proposed method shows positive R values, 0.25 for `Hey Snapdragon' and 0.40 for `Hey Snips'. If there is a common tendencies between positives and negatives across keywords, we can expect useful threshold decision rules from them. Here we tried a simple linear interpolation introduced in Section 2.3.

We search $\tau$ in Eq.(\ref{eq:tr1}) leveraging brute-force to get near 0.05 FAs per hour on average for 111 models. We set $\tau$ to 0.82 for baseline and 0.38 for the proposed method, and resulting FAs per hour are 0.049 for baseline and 0.050 for the proposed method on average.

Both method find the $\tau$ and reach target FAs per hour level, however, these two methods have dramatic difference in inter keywords. Inter keyword difference should be small in order to query-by-example system to work on any kind of keywords. For the baseline, `Hey Snapdragon' shows 0.001 FAs per hour while `Hey Snips' shows 0.088 FA per hour. Despite of using 6 to 7 times lower $B$, the proposed method shows exact same, 0.050 FAs per hour for both keyword `hey Snapdragon' and `Hey Snips'. Baseline shows 17.77 \% FRR at 6 dB noisy positives due to the low FAs per hour while the proposed method shows 3.95 \% FRR for 'hey Snapdragon'. The result is different from Table \ref{compare table}, because it uses given FAs per hour level for each model while this session use averaged FAs per hour.

\section{Conclusions}
\label{sec:typestyle}

In this paper, we suggest a simple and powerful approach for query-by-example on-device keyword spotting task. Our system uses user-specific queries, and CTC based AM outputs phonetic posteriorgram. We decode the output and build left-to-right FSTs as a hypothesis. The log likelihood is calculated as a score for testing. For on-device test, we suggest a method to predict a proper user and query specific threshold with the hypothesis. We generate query-specific negatives by shuffling the query in waveform. While many previous KWS approaches are not reproducible due to the limited data access, we tested our methods on public and well-known data. In the experiments, our approach showed promising and comparable performances to the latest predefined and query-by-example methods.
There is a limit to this work due to lack of public data, and we suggest naive approach for utilizing generated negatives. As a future work, we will study advanced way to predict threshold using the query-specific negatives, and test various keywords.

\vfill

\pagebreak

\bibliographystyle{IEEEbib}
\bibliography{strings,refs}

\begin{thebibliography}{10}

\bibitem{Guo2018}
J.~Guo, K.~Kumatani, M.~Sun, M.~Wu, A.~Raju, N.~Strom, and A.~Mandal,
\newblock ``Time-delayed bottleneck highway networks using a dft feature for
  keyword spotting,''
\newblock in {\em ICASSP, IEEE International Conference on Acoustics, Speech
  and Signal Processing - Proceedings}. IEEE, 2018, pp. 5489--5493.

\bibitem{Chen2018}
M.~Chen, S.~Zhang, M.~Lei, Y.~Liu, H.~Yao, and J.~Gao,
\newblock ``Compact feedforward sequential memory networks for small-footprint
  keyword spotting,''
\newblock in {\em {INTERSPEECH} 2018 -- 19\textsuperscript{th} Annual
  Conference of the International Speech Communication Association}, 2018, pp.
  2663--2667.

\bibitem{Shan2018}
C.~Shan, J.~Zhang, Y.~Wang, and L.~Xie,
\newblock ``Attention-based end-to-end models for small-footprint keyword
  spotting,''
\newblock in {\em {INTERSPEECH} 2018 -- 19\textsuperscript{th} Annual
  Conference of the International Speech Communication Association}, 2018, pp.
  2037--2041.

\bibitem{wangadversarial}
X.~Wang, S.~Sun, C.~Shan, J.~Hou, L.~Xie, S.~Li, and X.~Lei,
\newblock ``Adversarial examples for improving end-to-end attention-based
  small-footprint keyword spotting,''
\newblock in {\em ICASSP, IEEE International Conference on Acoustics, Speech
  and Signal Processing - Proceedings}. IEEE, 2019.

\bibitem{coucke2018}
A.~Coucke, M.~Chlieh, T.~Gisselbrecht, D.~Leroy, M.~Poumeyrol, and T.~Lavril,
\newblock ``Efficient keyword spotting using dilated convolutions and gating,''
\newblock in {\em arXiv preprint arXiv:1811.07684}, 2018.

\bibitem{Alvarez}
A.~Raziel and H.~Park,
\newblock ``End-to-end streaming keyword spotting,''
\newblock in {\em arXiv preprint arXiv: 1812.02802}, 2019.

\bibitem{hazen2009query}
T.J. Hazen, W.~Shen, and C.~White,
\newblock ``Query-by-example spoken term detection using phonetic posteriorgram
  templates,''
\newblock in {\em 2009 IEEE Workshop on Automatic Speech Recognition \&
  Understanding}. IEEE, 2009, pp. 421--426.

\bibitem{zhang2009unsupervised}
Y.~Zhang and J.R. Glass,
\newblock ``Unsupervised spoken keyword spotting via segmental dtw on gaussian
  posteriorgrams,''
\newblock in {\em 2009 IEEE Workshop on Automatic Speech Recognition \&
  Understanding}. IEEE, 2009, pp. 398--403.

\bibitem{anguera2013memory}
X.~Anguera and M.~Ferrarons,
\newblock ``Memory efficient subsequence dtw for query-by-example spoken term
  detection,''
\newblock in {\em 2013 IEEE International Conference on Multimedia and Expo
  (ICME)}. IEEE, 2013, pp. 1--6.

\bibitem{zhuang2016unrestricted}
Y.~Zhuang, X.~Chang, Y.~Qian, and K.~Yu,
\newblock ``Unrestricted vocabulary keyword spotting using lstm-ctc.,''
\newblock in {\em Interspeech}, 2016, pp. 938--942.

\bibitem{lugosch2018donut}
Loren Lugosch, Samuel Myer, and Vikrant~Singh Tomar,
\newblock ``Donut: Ctc-based query-by-example keyword spotting,''
\newblock {\em arXiv preprint arXiv:1811.10736}, 2018.

\bibitem{Chen2015}
G.~Chen, C.~Parada, and T.N. Sainath,
\newblock ``Query-by-example keyword spotting using long short-term memory
  networks,''
\newblock in {\em ICASSP, IEEE International Conference on Acoustics, Speech
  and Signal Processing - Proceedings}. IEEE, 2015, pp. 5236--5240.

\bibitem{he2017streaming}
Y.~He, R.~Prabhavalkar, K.~Rao, W.~Li, A.~Bakhtin, and I.~McGraw,
\newblock ``Streaming small-footprint keyword spotting using
  sequence-to-sequence models,''
\newblock in {\em 2017 IEEE Automatic Speech Recognition and Understanding
  Workshop (ASRU)}. IEEE, 2017, pp. 474--481.

\bibitem{audhkhasi2017end}
K.~Audhkhasi, A.~Rosenberg, A.~Sethy, B.~Ramabhadran, and B.~Kingsbury,
\newblock ``End-to-end asr-free keyword search from speech,''
\newblock {\em IEEE Journal of Selected Topics in Signal Processing}, vol. 11,
  no. 8, pp. 1351--1359, 2017.

\bibitem{Myer2018}
S.~Myer and V.S. Tomar,
\newblock ``Efficient keyword spotting using time delay neural networks,''
\newblock in {\em {INTERSPEECH} 2018 -- 19\textsuperscript{th} Annual
  Conference of the International Speech Communication Association}, 2018, pp.
  1264--1268.

\bibitem{Tang2018}
R.~Tang and J.~Lin,
\newblock ``Deep residual learning for small-footprint keyword spotting,''
\newblock in {\em ICASSP, IEEE International Conference on Acoustics, Speech
  and Signal Processing - Proceedings}. IEEE, 2018, pp. 5484--5488.

\bibitem{Pandey2018}
L.~Pandey and K.~Nathwani,
\newblock ``Lstm based attentive fusion of spectral and prosodic information
  for keyword spotting in hindi language,''
\newblock in {\em {INTERSPEECH} 2018 -- 19\textsuperscript{th} Annual
  Conference of the International Speech Communication Association}, 2018, pp.
  112--116.

\bibitem{fernandez2007application}
S.~Fern{\'a}ndez, A.~Graves, and J.~Schmidhuber,
\newblock ``An application of recurrent neural networks to discriminative
  keyword spotting,''
\newblock in {\em International Conference on Artificial Neural Networks}.
  Springer, 2007, pp. 220--229.

\bibitem{Menon2018}
R.~Menon, H.~Kamper, J.~Quinn, and T.~Niesler,
\newblock ``Fast asr-free and almost zero-resource keyword spotting using dtw
  and cnns for humanitarian monitoring,''
\newblock in {\em {INTERSPEECH} 2018 -- 19\textsuperscript{th} Annual
  Conference of the International Speech Communication Association}, 2018, pp.
  2608--2612.

\bibitem{Chen2018-2}
Y.~Chen, T.~Ko, L.~Shang, X.~Chen, X.~Jiang, and Q.~Li,
\newblock ``Meta learning for few-shot keyword spotting,''
\newblock in {\em arXiv preprint arXiv: 1812.10233}, 2018.

\bibitem{gravesCTC}
A.~Graves, S.~Fern{\'a}ndez, F.~Gomez, and J.~Schmidhuber,
\newblock ``Connectionist temporal classification: labelling unsegmented
  sequence data with recurrent neural networks,''
\newblock in {\em Proceedings of the 23rd international conference on Machine
  learning}. ACM, 2006, pp. 369--376.

\bibitem{paul1992design}
D.B. Paul and J.M. Baker,
\newblock ``The design for the wall street journal-based csr corpus,''
\newblock in {\em Proceedings of the workshop on Speech and Natural Language}.
  Association for Computational Linguistics, 1992, pp. 357--362.

\bibitem{panayotov2015librispeech}
V.~Panayotov, G.~Chen, D.~Povey, and S.~Khudanpur,
\newblock ``Librispeech: an asr corpus based on public domain audio books,''
\newblock in {\em 2015 IEEE International Conference on Acoustics, Speech and
  Signal Processing (ICASSP)}. IEEE, 2015, pp. 5206--5210.

\bibitem{Wang}
Y.~Wang, P.~Getreuer, T.~Hughes, R.F. Lyon, and R.A. Saurous,
\newblock ``Trainable frontend for robust and far-field keyword spotting,''
\newblock in {\em ICASSP, IEEE International Conference on Acoustics, Speech
  and Signal Processing - Proceedings}. IEEE, 2017, pp. 5670--5674.

\bibitem{muller2007dynamic}
M.~M{\"u}ller,
\newblock ``Dynamic time warping,''
\newblock {\em Information retrieval for music and motion}, pp. 69--84, 2007.

\end{thebibliography}

\end{document}